\documentclass[10pt,twocolumn,letterpaper]{article}

\usepackage{cvpr}              
\usepackage{times}
\usepackage{helvet}
\usepackage{courier}
\usepackage{graphicx}
\usepackage{amsmath}
\usepackage{diagbox}
\usepackage{amsmath}
\usepackage{amsfonts}
\usepackage{tabularx}  
\usepackage{array}     

 \usepackage{color}
\usepackage{url}
\usepackage{verbatim}
\usepackage{diagbox}
 \usepackage{color}
\usepackage{amssymb}

\usepackage{caption}
\usepackage{multirow}
\usepackage{booktabs}
\definecolor{cvprblue}{rgb}{0.21,0.49,0.74}
\usepackage[pagebackref,breaklinks,colorlinks,allcolors=cvprblue]{hyperref}

\def\paperID{9719} 
\def\confName{CVPR}
\def\confYear{2026}

\title{R4-CGQA: Retrieval-based Vision Language Models for Computer Graphics Image Quality Assessment}

\author{Zhuangzi Li, Jian Jin, Shilv Cai, Weisi Lin\\
College of Computing and Data Science, Nanyang Technological University, Singapore\\
}

\begin{document}
\maketitle
\begin{abstract}
Immersive Computer Graphics (CGs) rendering has become ubiquitous in modern daily life. However, comprehensively evaluating CG quality remains challenging for two reasons: (1) existing CG datasets lack systematic descriptions of rendering quality; and (2) existing CG quality assessment methods cannot provide reasonable text-based explanations. To address these issues, we first identify six key perceptual dimensions of CG quality from the user perspective and construct a dataset of $3.5\mathrm{K}$ CG images with corresponding quality descriptions. Each description covers CG style, content, and perceived quality along the selected dimensions. Furthermore, we use a subset of the dataset to build several question–answer benchmarks based on the descriptions in order to evaluate the responses of existing Vision Language Models (VLMs). We find that current VLMs are not sufficiently accurate in judging fine-grained CG quality, but that descriptions of visually similar images can significantly improve a VLM’s understanding of a given CG image. Motivated by this observation, we adopt retrieval-augmented generation and propose a two-stream retrieval framework that effectively enhances the CG quality assessment capabilities of VLMs. Experiments on several representative VLMs demonstrate that our method substantially improves their performance on CG quality assessment. The public dataset and code are at: {\color [RGB] {243,100,136} \textsl{https://github.com/lizhuangzi/R4-CGQA}}.
\end{abstract}

\begin{figure}[t]
  \centering
  \includegraphics[width=1.0 \linewidth]{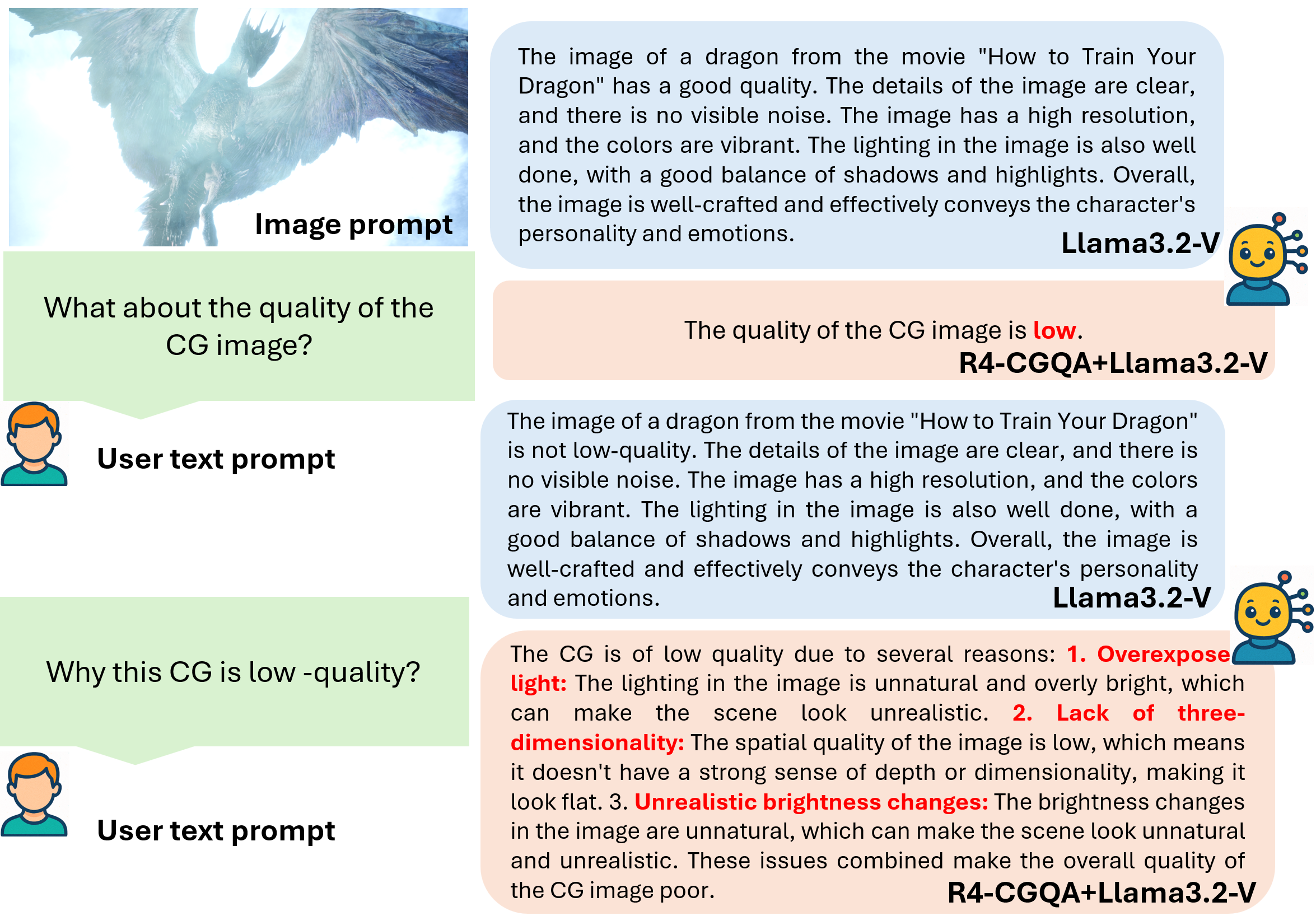}
  \caption{An example illustrating the capability of the proposed R4-CGQA. R4-CGQA not only makes the responses of large models more concise and accurate, but also effectively unleashes their potential on the CGQA task. \label{fig:Introduction0}}
\end{figure}

\section{Introduction}
High-quality computer graphics (CG) rendering is a crucial technology for many industrial applications, such as game design \cite{CG_game}, 3D animation \cite{CG_3Danmi}, and movie effects \cite{CG_Film}. In these scenarios, high-quality visual effects and immersive experiences have become essential to meet user expectations. Consequently, the industrial community requires intelligent quality assessment algorithms to guide the design and rendering of CG content. Although many image and AI-generated image quality assessment methods have been proposed recently, few works specifically consider the quality of CG images. Directly applying natural image quality assessment methods to CG images is inappropriate \cite{DBLP:conf/mmsp/WangZSMLZ22}, because CG images are entirely constructed by simulation including objects, textures, light sources, and camera perspectives and thus differ substantially from natural images in both distortions and perceptual characteristics.

\begin{figure}[t]
  \centering
  \includegraphics[width=1.0 \linewidth]{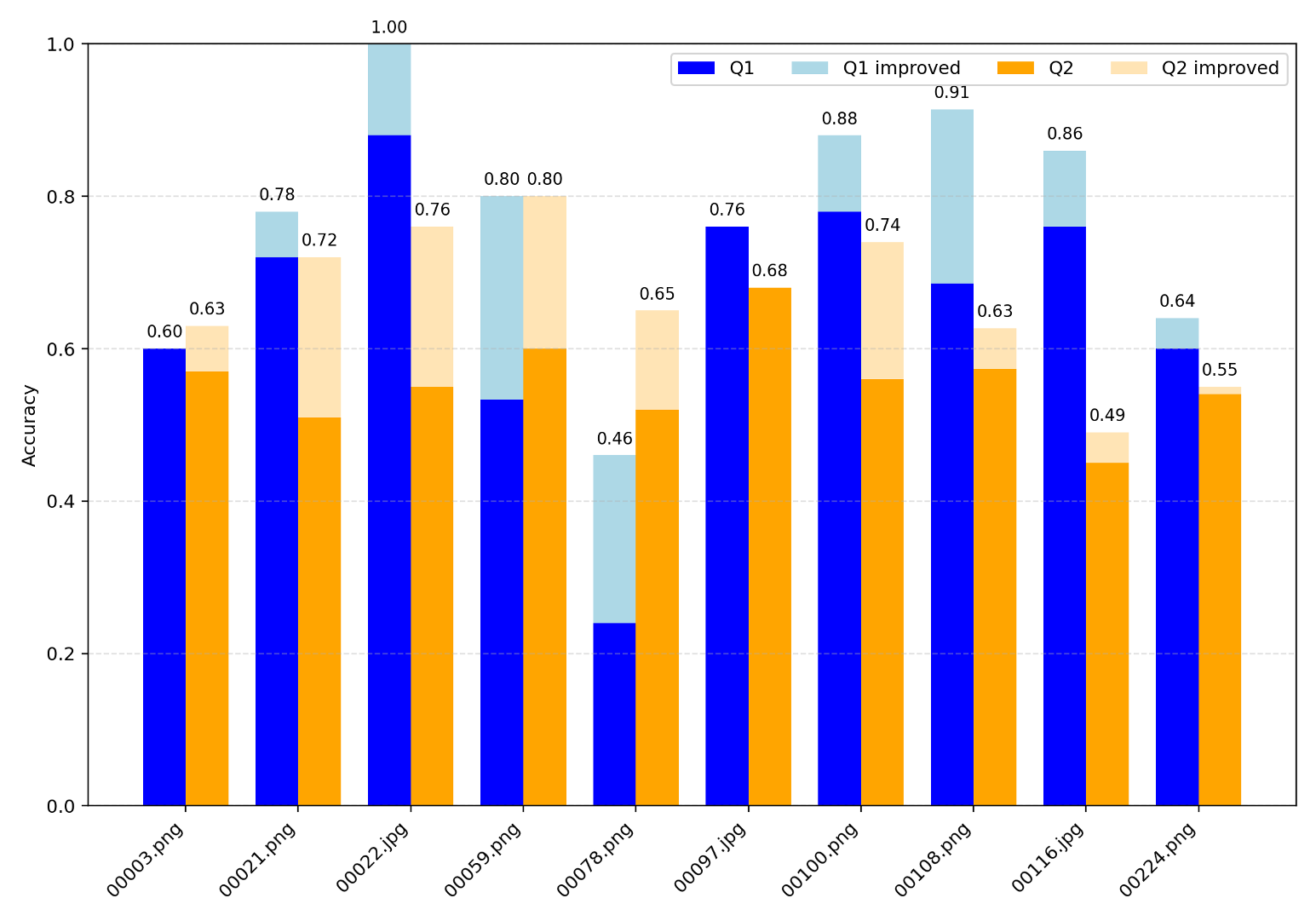}
  \caption{We prepare 10 CG images with associated questions and answers (Q1: multiple-choice questions and Q2: yes-or-no questions) and evaluate answer accuracy on LLaVA-13B \cite{DBLP:conf/nips/LiuLWL23a}. Dark bars indicate accuracy when the evaluation is directly provided by the VLM; light bars indicate accuracy when reference descriptions from visually similar CG images are provided to the VLM. \label{fig:Introduction}}
\end{figure}

To address this problem, several works have established CG image databases to provide discriminative standards for CG quality assessment (CGQA). Specifically, Wang et al. \cite{DBLP:conf/mmsp/WangZSMLZ22} collected $1.2\mathrm{K}$ CG images from 3D movies and 3D games with human scores, and show that deep learning based methods are more consistent with human perception than approaches based on hand-crafted features. Zhang et al. \cite{CGIQA_6k} constructed a CG dataset containing about $6\mathrm{K}$ images with scores and proposed a no-reference, deep learning–based CGQA model that simultaneously exploits distortion and aesthetic quality representations. However, these databases only contain subjective quality scores; in other words, their evaluation perspective is inevitably restricted. We argue that a good CGQA method should reveal the reasons behind human judgments and further provide clear guidance for anticipated CG rendering. Therefore, we construct a new dataset containing $3.5\mathrm{K}$ CG images with corresponding text-based quality descriptions. We ask professional participants to describe the three most salient items from six dimensions including lighting, material, color, atmosphere, realism, and space. This dataset enables language models to learn more accurate and comprehensive CGQA capabilities.

In recent years, Vision Language Models (VLMs) have exhibited impressive capabilities in quality description and reason\cite{DBLP:journals/corr/abs-2405-18842}. Chen et al. \cite{DBLP:journals/vciba/ChenHNCLSW24} presented an image QA system that combines VLMs with ChatGPT. Zhu et al. \cite{DBLP:conf/nips/Zhu0LZC0FZL024} proposed an adaptive image QA method by teaching a VLM to perform comparative judgments. The Q-series works \cite{DBLP:conf/iclr/0001Z0CLWLSYZL24,DBLP:conf/cvpr/0001Z0CLWXLHZXS24,DBLP:conf/icmcs/ZhangWJLZSLMSJLZ24,lu2025q} propose benchmarks, tuning methods, and evaluation approaches for various degraded images. Other works \cite{DBLP:conf/mm/Wang0ZJJZMZ24,DBLP:conf/aaai/TianLCZ0K25} focus on evaluating the quality of AI-generated images. However, VLMs are prone to hallucinations due to uncertainty in some knowledge domains, especially in the field of CGQA. Fine-tuning a VLM requires substantial computational resources and large amounts of training data, and it is not conducive to keeping the model’s knowledge up to date.

Our preliminary study shows that quality descriptions of similar CG images can provide informative knowledge for understanding the quality of target CG images. Specifically, as shown in Figure~\ref{fig:Introduction}, we select 10 CG images and, for each, provide descriptions of visually similar examples to a VLM, which is then asked to answer multiple-choice questions (Q1) and yes-or-no questions (Q2). The chart clearly shows that incorporating descriptions from CG images that are visually similar to the target CG generally enhances the accuracy of VLMs in answering related questions. Besides, we notice that incorporating descriptions irrelevant to target images is detrimental to the VLM’s understanding. These observations motivate us to utilize retrieval-based technology for CGQA based on our CG database. However, defining similarity between two CG images is challenging. In this paper, we jointly consider content similarity and quality similarity in the retrieval process, ensuring that the example descriptions are appropriate. We conduct experiments on several VLMs, showing consistent improvements brought by our approach. Our method is general and can be applied into many VLMs, e.g., Llama3.2-V, as shown in Figure~\ref{fig:Introduction0}. Our contributions are summarized as follows:

\begin{itemize}
\item We propose a brand new CG quality assessment dataset with descriptions that consider six dimensions of CG quality, providing a suitable data foundation for CGQA. To the best of our knowledge, it is the first dataset specifically designed to systematically explain the quality of CG images.
\item Based on Bayesian theory, we propose a novel and general CG quality assessment framework for existing VLMs that incorporates both content similarity and quality similarity in the retrieval phase.
\item We provide testing benchmarks based on our dataset, and conduct experiments on several representative VLMs including LLaVA, Llama 3.2-Vision, and Qwen2.5-VL, which demonstrate the effectiveness of our approach. 
\end{itemize}

\section{Related Work}
\subsection{Conventional IQA}
Image quality assessment (IQA) methods are commonly categorized according to the availability of a reference image: full-reference (FR) \cite{DBLP:conf/hvei/WangS05,DBLP:journals/tip/SheikhB06}, reduced-reference (RR) \cite{DBLP:conf/cvpr/ZhangIESW18}, and no-reference (NR, also called blind IQA) \cite{DBLP:journals/tip/WangBSS04,DBLP:journals/tip/BosseMMWS18,DBLP:journals/tip/MaLZDWZ18,DBLP:conf/cvpr/SuYZZGSZ20}. By the 2010s, the IQA community had established this FR/RR/NR taxonomy together with a suite of engineered algorithms (e.g., SSIM, FSIM, VIF, GMSD, BLIINDS, BRISQUE) that significantly improved correlation with human judgments over raw error measures \cite{DBLP:conf/icip/YouK21}. Although these hand-crafted techniques are effective for conventional distortions, their performance tends to plateau on complex, real-world images.
Deep learning–based IQA leverages large-scale annotated data to train networks that directly predict perceptual quality scores.
Kim et al. \cite{BIECON} proposed the BIECON model, which uses patch-level CNN features to predict local quality scores and then aggregates them into a global subjective score. 
Su et al. \cite{DBLP:journals/tip/BosseMMWS18} designed a self-adaptive hyper-network architecture for blind IQA in the wild and decomposed the IQA pipeline into three stages: content understanding, perceptual rule learning, and quality prediction. 
Varga \cite{DBLP:journals/corr/abs-2011-05139} employed Inception modules to capture multi-scale texture and structure, reflecting the need to handle distortions at various frequencies and patch sizes simultaneously.

Inspired by the success of transformers in vision tasks, researchers have further applied self-attention networks to IQA. You et al. \cite{TIQA,DBLP:conf/cvpr/CheonYKL21} introduced TRIQ, a transformer-based NR-IQA model that appends a shallow transformer encoder to CNN features. TRIQ’s adaptive positional encoding enables the processing of images with arbitrary resolution, and its multi-head attention mechanism captures global image context more effectively than conventional CNNs. In parallel, Ke et al. \cite{DBLP:conf/iccv/KeWWMY21} proposed MUSIQ, a multi-scale image quality transformer that processes full-resolution images by combining patch embeddings across multiple scales. Transformers have also been deployed for FR-IQA: Cheon et al. \cite{DBLP:conf/cvpr/CheonYKL21} developed Image Quality Transformer (IQT), which uses a Siamese encoder–decoder transformer to compare a reference image with its distorted counterpart. 
Alongside these novel architectures, researchers have addressed data limitations through semi-supervised learning and synthetic data generation. For example, Cao et al. \cite{DBLP:conf/cvpr/CaoWRYZ22} combined positive–unlabeled learning with a dual-branch network to exploit unlabeled degraded images for FR-IQA, employing the sliced Wasserstein distance to handle spatial misalignment. Overall, recent work has established deep learning–based IQA as the dominant paradigm: CNN- and transformer-based models now achieve near human-level correlations on many benchmarks. The incorporation of attention mechanisms, multi-task learning (e.g., joint distortion identification), and large-scale training has substantially improved the generalization of IQA models across distortion types and content domains. However, these methods generally do not provide specific, human-interpretable reasons for their predicted scores, which limits their ability to guide subsequent modifications to the evaluated images.

\subsection{IQA with Large Vision Language Models}
Most recently, large vision language models (VLMs) such as BLIP-2 \cite{DBLP:conf/icml/0008LSH23}, GPT-4 \cite{gpt4}, and Qwen2.5-VL \cite{DBLP:journals/corr/abs-2502-13923} have emerged, learning rich vision–language correlations from large-scale image-text pairs. IQA research is increasingly influenced by the rise of these VLMs. Instead of training dedicated IQA networks from scratch, researchers are exploiting foundation models trained on massive image or image–text datasets to improve perceptual quality assessment. Wang et al. \cite{DBLP:conf/aaai/WangCL23} explored CLIP for zero-shot IQA, hypothesizing that CLIP’s learned visual features and alignment with human descriptive language can serve as a proxy for human quality judgments. They demonstrated that, by crafting appropriate text prompts (e.g., “a high-quality photo” vs. “a low-quality photo”) and measuring an image’s similarity to these descriptions in CLIP’s embedding space, one can obtain quality predictions without task-specific training. 
Chen et al. introduced IQAGPT \cite{DBLP:journals/vciba/ChenHNCLSW24}, an innovative computed tomography IQA system that integrates an image-quality captioning VLM with ChatGPT to generate both quality scores and textual reports. Although these VLMs were not originally designed for conventional IQA, they can complement quality assessment by checking aspects such as whether an image contains unrealistic objects or mismatched captions \cite{wang2024understandingevaluatinghumanpreferences}. Zhu et al. \cite{DBLP:conf/nips/Zhu0LZC0FZL024} formulated IQA as a comparative reasoning task suitable for large multimodal models (LMMs), training a vision–language model to make pairwise quality comparisons between images by generating instructional prompts during training (e.g., “Which image is better?” given two inputs). However, these approaches still exhibit limited explainability of their scores and therefore provide weak guidance for improving image quality.

   \begin{figure}[t]
  \centering
  \includegraphics[width=1.0 \linewidth]{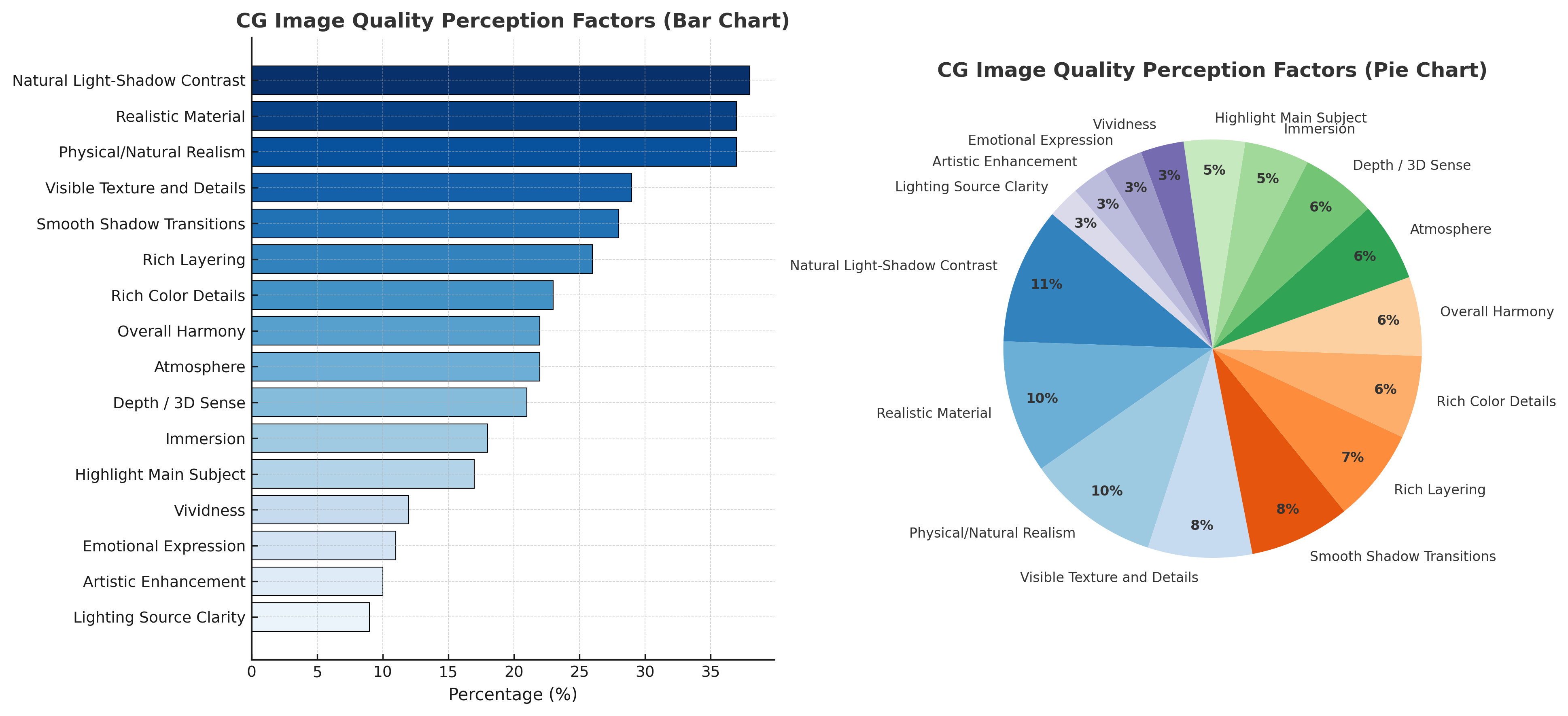}
  \caption{CG Image Quality Perception Factors. \label{fig:Prestatistics} }
\end{figure}

   \begin{figure*}[t]
  \centering
  \includegraphics[width=1.0 \linewidth]{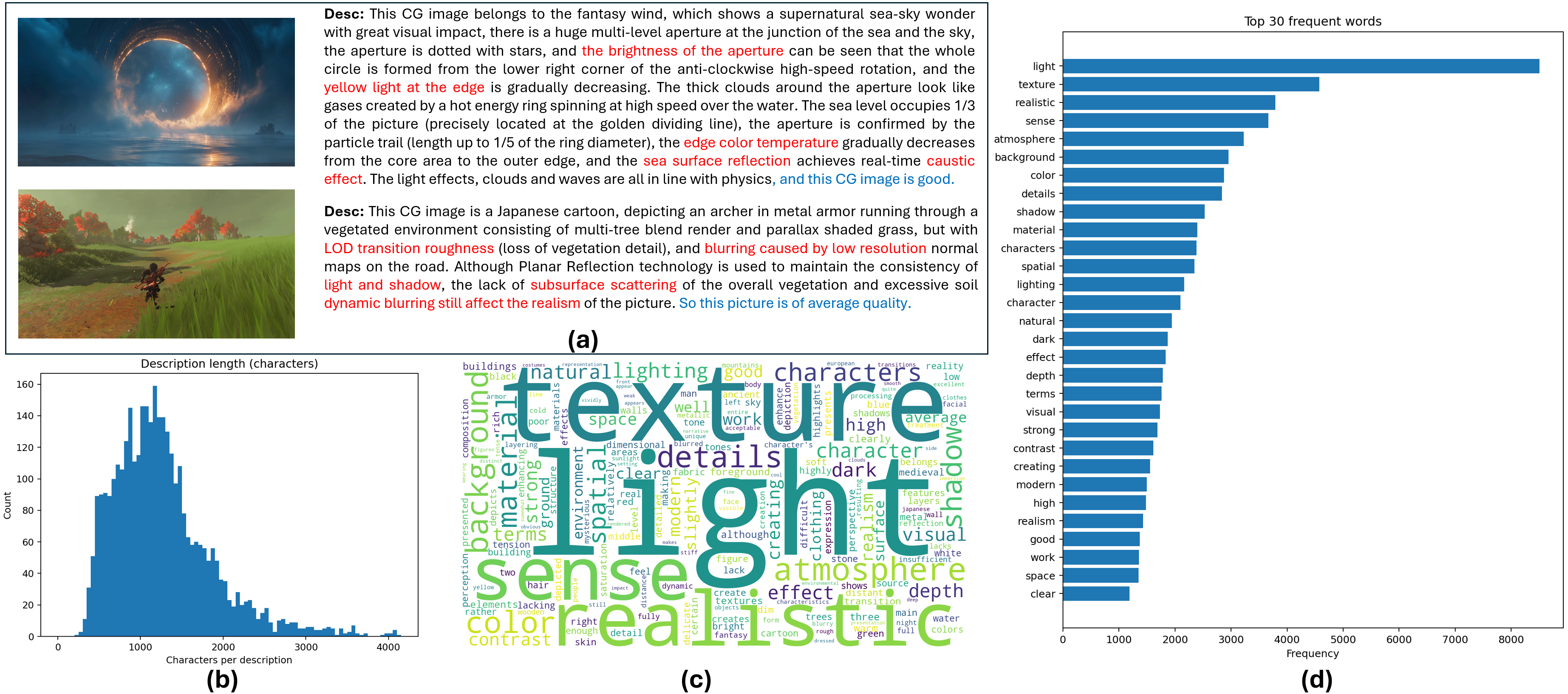}
  \caption{Overview of dataset characteristics.
(a) Example image–description pairs illustrating diverse visual content.
(b) Distribution of description lengths across the dataset.
(c) Word cloud visualization of the most frequent descriptive terms, highlighting key perceptual attributes such as texture, light, and realism.
(d) Top 30 most frequent words and their corresponding frequencies. \label{fig:dataset} }
\end{figure*}

Q-Bench \cite{DBLP:conf/iclr/0001Z0CLWLSYZL24} presents a holistic benchmark that systematically evaluates the capabilities of multimodal large language models (MLLMs) in three realms: low-level visual perception, low-level visual description, and overall visual quality assessment. Building on this, Wu et al. proposed Q-Instruct \cite{DBLP:conf/cvpr/0001Z0CLWXLHZXS24}, which uses image quality descriptions and ChatGPT to construct large-scale question–answer pairs for fine-tuning VLMs. Q-Adapt \cite{lu2025q} aims to reduce conflicts and achieve synergy between overall quality explanations and attribute-wise perceptual judgments. Zhang et al. \cite{DBLP:conf/icmcs/ZhangWJLZSLMSJLZ24} proposed Q-Boost to enhance VLM performance on low-level IQA tasks in a boosting manner.
In parallel, recent studies on AI-generated images have expanded the notion of “image quality” to include authenticity and semantic alignment in addition to distortion fidelity \cite{DBLP:conf/mm/Wang0ZJJZMZ24,AzizRehmanDanishGrolinger2025GLIPS}. Correspondingly, several AI-generated image quality assessment datasets for VLMs have been proposed, such as AIGIQA2K \cite{Li2024AIGIQA20K} and AIGI-VC \cite{DBLP:conf/aaai/TianLCZ0K25}. These models and datasets provide powerful high-level representations and semantic understanding, helping IQA move beyond low-level distortion measurement toward holistic perceptual assessment. However, in the computer graphics (CG) domain, there is still a lack of CG datasets specifically designed for VLM-based quality assessment; some existing datasets provide only scalar quality scores \cite{DBLP:conf/mmsp/WangZSMLZ22,CGIQA_6k} and omit a detailed analysis of the quality dimensions of CG images.

\section{Our CGQA Dataset}
In recent years, the use of computer-generated (CG) images has become increasingly widespread in areas such as gaming, virtual reality, film production, and digital content creation. With the rapid development of these industries, there is a growing need for comprehensive and well-structured CG image datasets to support research in image quality assessment, realism analysis, and visual perception. Several CG datasets have been proposed; however, many of them are of limited quality, as most images are low-resolution. Moreover, they typically provide only mean opinion score (MOS) labels for CG images, without annotating detailed perceptual dimensions or describing CG quality. Such datasets are not well suited to today’s intelligent evaluation systems based on multimodal large models. Existing databases \cite{CGIQA_6k} provide CG images with score labels, but they do not offer reasons or explanations for the assigned scores. We argue that evaluating the quality of CG images requires considering many aspects, such as lighting, color, and spatial composition. However, given a CG image, it is non-trivial to produce a consistent and reasonable textual description. 

To address this problem, we consulted CG industry practitioners and summarized the key terms they focus on, as shown in Figure~\ref{fig:Prestatistics}. 
Based on this analysis, we abstract six key dimensions for evaluating CG quality: lighting quality, material quality, color quality, atmosphere, realism, and space. We then recruited 15 participants with gaming experience or professional backgrounds in CG to construct the dataset. Each participant underwent relevant training before evaluation to ensure a consistent rating scale and a unified perspective on description. During annotation, we asked them to describe the CG quality from at least three salient dimensions and to provide an overall quality conclusion based on their descriptions. 
The CG image dataset discussed in this work comprises high-resolution images generated using a variety of rendering engines and techniques. In total, it contains $3.5\mathrm{K}$ CG images with diverse styles, including medieval realism, modern realism, dark realism, fantasy, and cartoons etc.. The CG images are collected from Wallpaper Engine, some high-definition game CG screenshots (World of Warships, World War Z, World of Tanks, Elden Ring, etc.), CGIQA-6K \cite{CGIQA_6k} (images with MOS scores greater than 3.2 and a subset of images with scores less than 3.2 selected as low-quality CG candidates), and several CG packages purchased online. The resolutions range from 1080p to 4K. To ensure annotation quality, the images were divided into subsets and distributed among annotators so that each annotator viewed a diverse selection of content. 

Figure~\ref{fig:dataset} illustrates representative samples and statistical characteristics of our dataset.
Panel (a) shows two examples of annotated CG images along with their detailed textual descriptions.
In each description, the red-highlighted phrases correspond to fine-grained attributes focusing on different perceptual dimensions, such as lighting effects, color temperature, reflections, textures, or material realism; these segments emphasize which specific aspects are being described.
The blue-highlighted phrases summarize the overall quality assessment, indicating the perceived realism or aesthetic judgment of the image (e.g., “this CG image is good” or “of average quality”).
Panel (b) presents the distribution of description lengths; most descriptions are longer than $1{,}000$ characters, indicating that the majority of entries contain rich detail.
Panel (c) visualizes the most frequent descriptive terms via a word cloud, where high-frequency words such as texture, light, and realistic highlight the dataset’s strong focus on visual perception attributes.
Panel (d) lists the top 30 most frequent words and their occurrence counts, further quantifying the dominant vocabulary patterns in the corpus.

\begin{figure}[t]
  \centering
  \includegraphics[width=1.0 \linewidth]{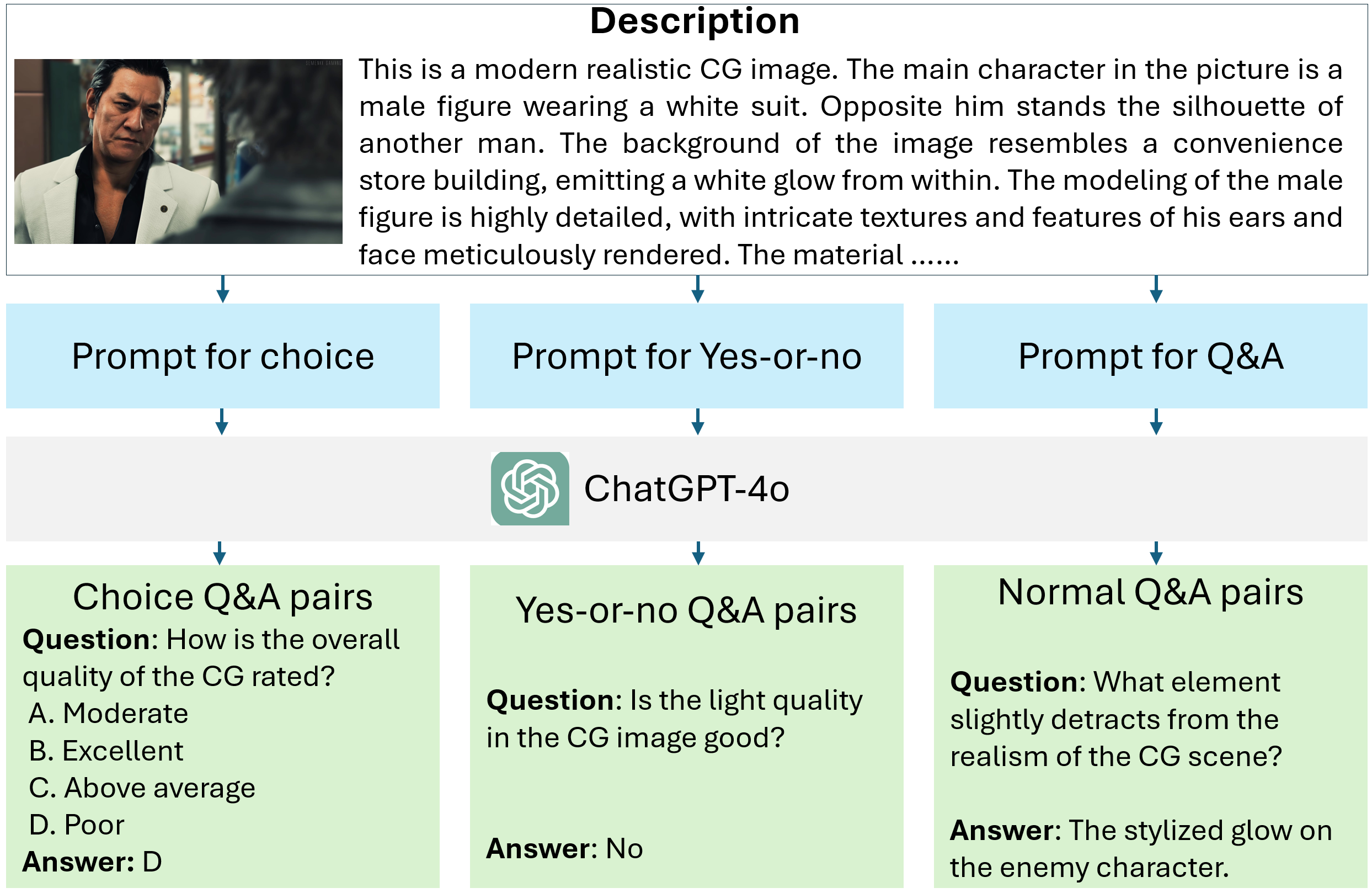}
  \caption{GPT-4o is used to generate three types of questions for the validation and testing sets. \label{fig:testingset}}
\end{figure}

To facilitate the use of our dataset, we split it into three parts: a base set, a validation set, and a testing set. The base set contains $3{,}190$ CG images and can be used for training, fine-tuning, and database construction. The validation and testing sets include $90$ and $220$ images, respectively. For each image in the validation and testing sets, we use GPT-4o to generate three types of questions: multiple-choice, yes-or-no, and normal Q\&A, as shown in Figure ~\ref{fig:testingset}, and each type includes at least five questions. Therefore, the validation and testing sets together contain more than $5\mathrm{K}$ question-answer pairs. These are sufficient to serve as benchmarks for evaluating various VLMs.


  \begin{figure*}[t]
  \centering
  \includegraphics[width=0.9 \linewidth]{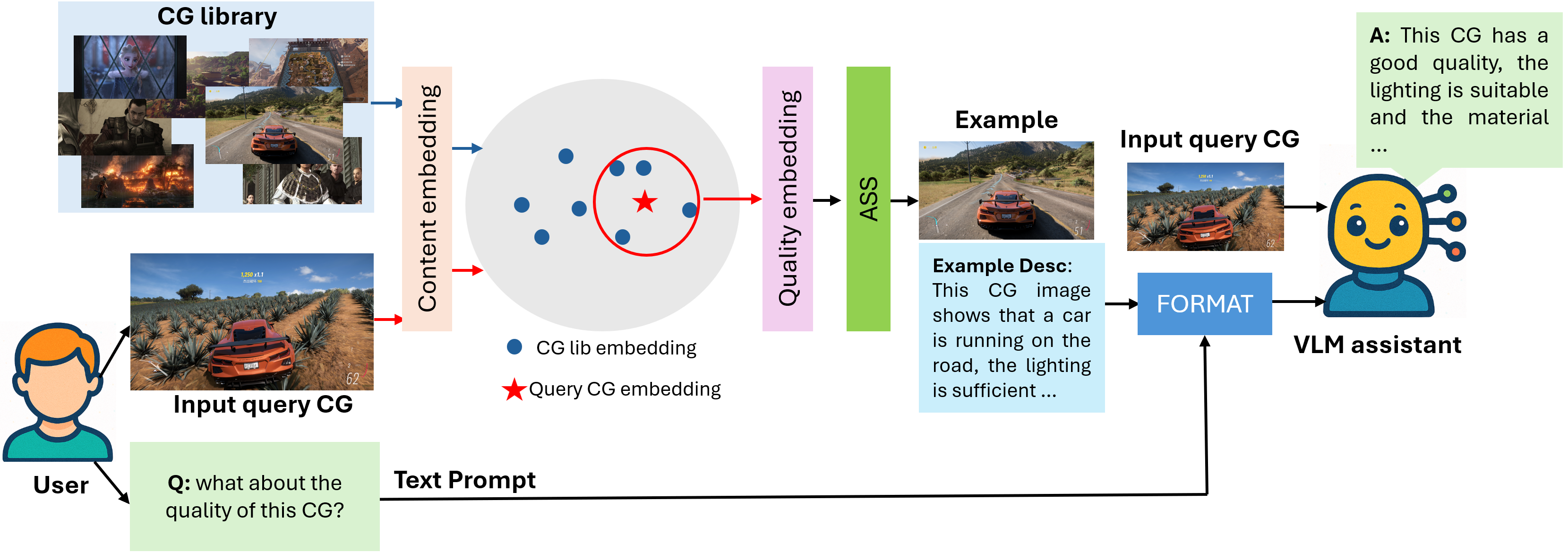}
  \caption{\label{fig:framework}  Overview of R4-CGQA. A query CG image and a CG library are embedded in a content space, and a local neighborhood around the query is formed. Within this neighborhood, quality embeddings are computed and combined to select the most similar example, whose textual description is used as additional context when querying the VLM assistant together with the input CG, producing a final quality judgement and explanation. The CG library is built by our \textbf{base set} and it is scalable; ASS denotes average similarity search.}
\end{figure*}
\section{Our R4-CGQA}
\subsection{Framework based on Bayesian theory}
We consider a retrieval-based method to empower CGQA ability for VLMs. Compared with traditional fine-tuning, this approach not only has lower performance overhead, but also offers greater flexibility when dealing with unseen new CG images. We attempt to tap into the potential of the large model to answer questions related to CG by referring to similar CG image information, which can be written as a Bayesian style. 
Let $x$ denote a query CG image, $q$ a natural-language question about this
image (e.g., ``What about the quality of this CG?''), and $a$ the answer
produced by the VLM.
We assume access to a CG library $\mathcal{D}=\{(x_i,t_i)\}_{i=1}^{N}$, 
where $x_i$ is a library image and $t_i$ is an associated human-written
description (e.g., quality comments or scene explanations).
At inference time, we select one library index $I\in\{1,\dots,N\}$, use
$t_I$ as an example description, and build a retrieval-augmented prompt
$\mathrm{Prompt}(x,q,t_I)$ for the VLM.

Given a fixed VLM with parameters $\phi$, the answer distribution is
\begin{equation}
a\sim p_{\phi}\big(a \mid x,q,\mathrm{Prompt}(x,q,t_I)\big).
\end{equation}
If we had a task-specific utility function $U(a)$ (e.g., combining
multiple-choice accuracy and textual quality scores), the ideal retrieval
policy would maximize the expected utility
\begin{equation}
I^\star
  = \arg\max_{i\in\{1,\dots,N\}}
      \mathbb{E}_{a\sim p_{\phi}(\cdot \mid x,q,\mathrm{Prompt}(x,q,t_i))}
      \big[U(a)\big].
\label{eq:ideal}
\end{equation}
However, directly optimizing~\eqref{eq:ideal} is intractable.
Instead, we adopt a Bayesian retrieval view and approximate this objective by
maximizing a proxy posterior
\begin{equation}
P(I=i\mid x,q)\propto P(z(x)\mid I=i)\,\pi_i,
\label{eq:posterior-bayes}
\end{equation}
where $z(x)$ denotes a latent representation of the query extracted from one
or more embedding spaces, and $\pi_i=P(I=i)$ is a prior over library
indices.
The retrieval rule becomes the maximum a posteriori (MAP) estimator
\begin{equation}
I^\star = \arg\max_{i}
          \log P\big(z(x)\mid I=i\big)
          + {\log \pi_i}.
\label{eq:map-generic}
\end{equation}

In practice we do not have an exact generative model $P(z(x)\mid I=i)$.
Instead, we assume that $z(x)$ consists of $M$ different views
$z^{(1)}(x),\dots,z^{(M)}(x)$ (e.g., content and quality embeddings), and we
define view-specific similarities $s_m(x,x_i)$ between the query and a
library image.
A simple and widely used approximation is to model each view likelihood
by an exponential family of the similarity,
\begin{equation}
P\big(z^{(m)}(x)\mid I=i\big)\propto
  \exp\!\big(\beta_m s_m(x,x_i)\big) ,
\label{eq:exp-sim}
\end{equation}
with non-negative weights $\beta_m$, where $m=1,\dots,M$.
Assuming conditional independence between views, the log-posterior then
takes the form
\begin{equation}
\log P(I=i\mid x,q)
  = {c}
    + \sum_{m=1}^{M} \beta_m s_m(x,x_i)
    + \log \pi_i,
\label{eq:log-posterior-generic}
\end{equation}
where $c$ denotes const, and the MAP rule~\eqref{eq:map-generic} reduces to a weighted similarity
fusion:
\begin{equation}
I^\star
   = \arg\max_{i}
      \Bigg[
        \sum_{m=1}^{M} \beta_m s_m(x,x_i)
        + \log \pi_i
      \Bigg].
\label{eq:map-sim}
\end{equation}
Our concrete system, described next, is an instantiation of this generic
MAP retrieval rule with two views ($M=2$): image content and quality, and we use a uniform prior, so that
$\log \pi_i$ becomes a constant and does not affect the equation.

\subsection{Content and quality based retrieval}
Our previous work consider using content similarity retrieval only, but images with the same content may have large differences in quality. Specifically, CLIP \cite{CLIP} is not sensitive to image degradation. If the quality differences between ``similar" images are significant, feeding them to VLMs may cause misleading. 
Therefore, based on Eq.~(\ref{eq:map-sim}), we introduce content and quality based retrieval by combining a CLIP-based content embedding \cite{CLIP} and a REIQA-based quality embedding \cite{ReIQA} for every image.
As shown in Figure~\ref{fig:framework}, for any image $z$, let $f_c(z)\in\mathbb{R}^{d_c}$ and $f_q(z)\in\mathbb{R}^{d_q}$ denote the content and quality embeddings, respectively.
We use $\ell_2$-normalized embeddings
\begin{equation}
\hat f_c(z)=\frac{f_c(z)}{\lVert f_c(z)\rVert_2},\qquad
\hat f_q(z)=\frac{f_q(z)}{\lVert f_q(z)\rVert_2}.
\end{equation}

\paragraph{Stage~1: CLIP-based content retrieval.}
For the query image $x$ and each library image $x_i$ we compute the cosine
similarity in the CLIP space,
\begin{equation}
s_c(x,x_i)=\hat f_c(x)^\top \hat f_c(x_i).
\label{eq:sim-clip}
\end{equation}
To obtain an efficient global search, all $f_c(x_i)$ are indexed by FAISS \cite{faiss2017}.
At inference time, we retrieve a small candidate set of $K$ nearest
neighbours
\begin{equation}
\mathcal{S}_K(x)
   = \operatorname*{TopK}_{i\in\{1,\dots,N\}} s_c(x,x_i),
\label{eq:topk}
\end{equation}
corresponding to the red circle around the query embedding in
Figure~\ref{fig:framework}.

\begin{figure}[t]
  \centering
 \includegraphics[width=0.95 \linewidth]{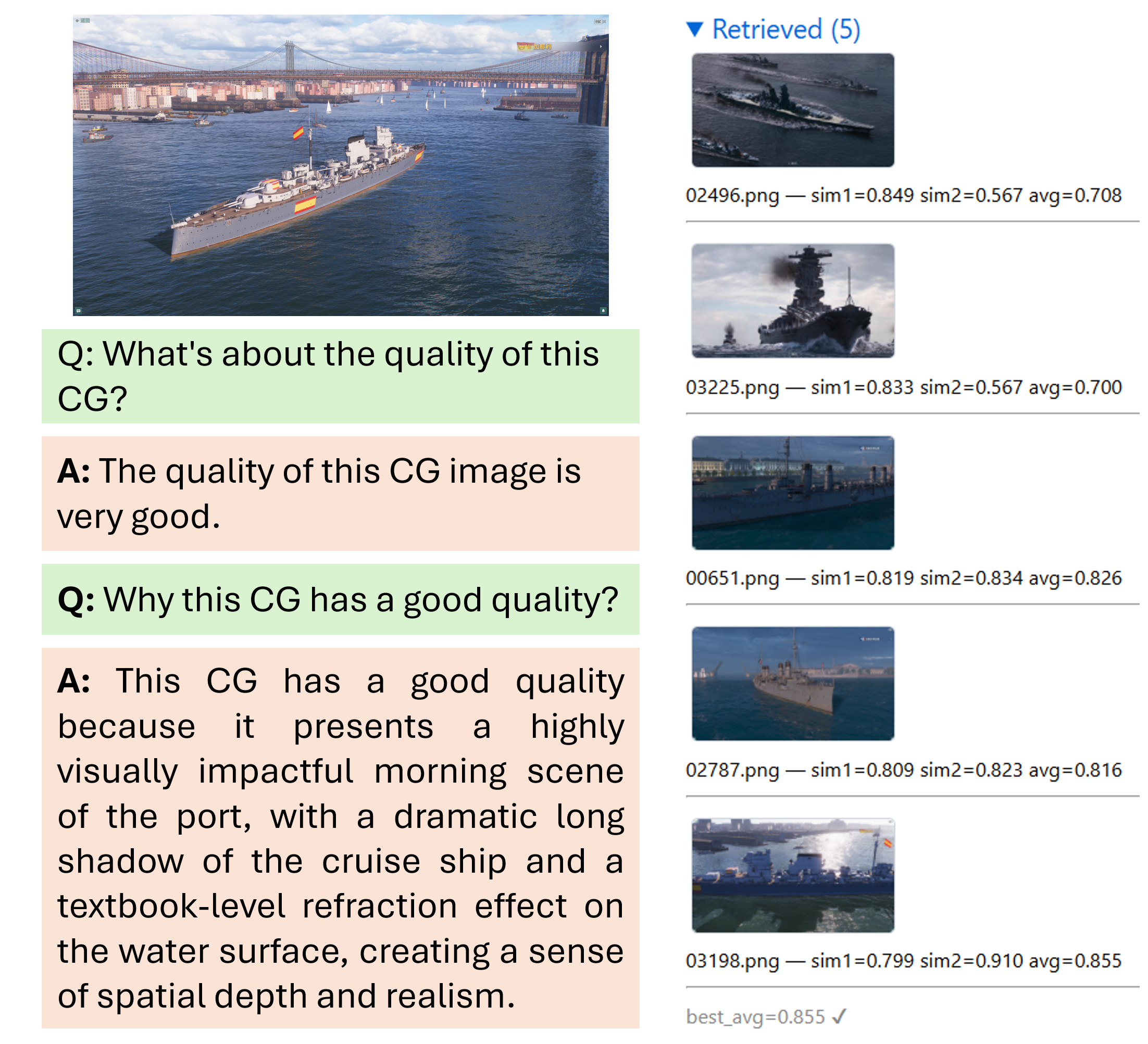}
  \caption{%
    Qualitative example of the R4-CGQA system. The left panel shows the query CG; the right panel
    shows the top-$5$ retrieved images with their CLIP similarity
    (\texttt{sim1}), REIQA-based similarity (\texttt{sim2}) and averaged
    score (\texttt{avg}).  The bottom text illustrates VLM answers to two
    questions, combining a scalar quality judgment with a detailed natural
    language explanation.}
  \label{fig:ass}
\end{figure}

\paragraph{Stage~2: quality embeddings and similarity fusion.}
For each candidate $i\in\mathcal{S}_K(x)$ we compute a quality embedding
$f_q(x_i)$ using the quality-aware ResNet backbone and the corresponding
cosine similarity
\begin{equation}
s_q(x,x_i)=\hat f_q(x)^\top \hat f_q(x_i).
\label{eq:sim-qual}
\end{equation}
We then fuse the content and quality similarities by a simple average,
\begin{equation}
S(x,x_i)
   = \frac{1}{2}\,s_c(x,x_i) + \frac{1}{2}\,s_q(x,x_i),
   \qquad i\in\mathcal{S}_K(x).
\label{eq:avg-sim}
\end{equation}
The index with the highest fused score is selected:
\begin{equation}
I^\star(x) = \arg\max_{i\in\mathcal{S}_K(x)} S(x,x_i).
\label{eq:argmax-inst}
\end{equation}
In practice we also apply a similarity threshold $\tau_{\mathrm{sim}}$:
if $\max_{i\in\mathcal{S}_K(x)} S(x,x_i) < \tau_{\mathrm{sim}}$, no example
description is used for this query and the VLM is prompted only with the
query image and question. An example of retrieval results and answers is shown in Figure.~\ref{fig:ass}.


\newcommand{\tabincell}[2]{
\begin{tabular}{@{}#1@{}}#2\end{tabular}
}

\begin{table*} [t] 
\begin{center}
\small
\caption{\label{fulltable}%
Evaluating state-of-the-art VLMs on our testing set, the ``Original" part denotes the results of directly using VLMs to answer questions, and the R4-CGQA denotes the results of our approach. Scores in \textcolor{red}{red} denote the improvement.}
\begin{tabular}{c|ccc|ccc|c}
\hline

\multirow{2}{*}{VLMs} & \multicolumn{3}{c|}{Original} & \multicolumn{3}{c|}{R4-CGQA} & \multirow{2}{*}{$\uparrow$ Average } \\ \cline{2-7} 
 & Choice & Yes-or-no & Q\&A  & Choice & Yes-or-no & Q\&A & \\ \hline \hline
{LLava-1.6-7B} & $51.70\%$ & $49.73\%$ & $2.20$  &  $58.06\%$ \tiny{\textcolor{red}{(+6.36\%)}} & $58.84\%$ \tiny{\textcolor{red}{(+9.11\%)}} & $2.50$ \tiny{\textcolor{red}{(+6.00\%)}} & {7.16\%} \\ \hline
{LLava-1.6-13B} & $53.96\%$ & $51.34\%$ &  $2.22$ &  $61.43\%$ \tiny{\textcolor{red}{(+7.47\%)}} & $60.37\%$ \tiny{\textcolor{red}{(+9.03\%)}} & $2.56$ \tiny{\textcolor{red}{(+6.80\%)}} & {7.77\%}  \\ \hline
{Llama 3.2-V-11B} & $64.59\%$ & $56.87\%$ & $1.93$ &  $67.28\%$ \tiny{\textcolor{red}{(+2.69\%)}} & $67.26\%$ \tiny{\textcolor{red}{(+10.39\%)}} & $2.31$ \tiny{\textcolor{red}{(+7.60\%)}}  & {6.89\%}\\ \hline
{Gemma3-4B} & $66.03\%$ & $53.55\%$ & $1.05$  &  $66.82\%$ \tiny{\textcolor{red}{(+0.79\%)}} & $65.22\%$ \tiny{\textcolor{red}{(+11.67\%)}} & $2.32$ \tiny{\textcolor{red}{(+25.40\%)}}  & {12.62\%}\\ \hline
{MiniCPM-V-8B} & $60.05\%$ & $53.47\%$ & $1.92$ &  $67.63\%$ \tiny{\textcolor{red}{(+7.58\%)}} & $61.98\%$ \tiny{\textcolor{red}{(+8.51\%)}} & $2.34$ \tiny{\textcolor{red}{(+8.40\%)}} & {8.16\%}\\ \hline
{Bakllava-7B} & $43.72\%$& $52.85\%$ & $1.67$  &  $55.97\%$ \tiny{\textcolor{red}{(+12.25\%)}}  & $61.17\%$ \tiny{\textcolor{red}{(+8.32\%)}}  & $1.96$ \tiny{\textcolor{red}{(+5.80\%)}} & {8.79\%} \\ \hline
{Pixtral-12B} & $70.11\%$& $66.20\%$ & $2.78$  &  $73.12\%$ \tiny{\textcolor{red}{(+3.01\%)}}  & $69.10\%$ \tiny{\textcolor{red}{(+2.90\%)}}  & $2.89$ \tiny{\textcolor{red}{(+2.20\%)}} & {2.70\%} \\ \hline
{LlavaNext-8B} & $69.41\%$ & $61.92\%$ & 2.68  &  $72.74\%$ \tiny{\textcolor{red}{(+3.33\%)}} &  $67.43\%$ \tiny{\textcolor{red}{(+5.51\%)}} & $2.72$ \tiny{\textcolor{red}{(+0.80\%)}} & {3.21\%} \\ \hline
{LlavaNext-32B} & $75.00\%$ & $63.21\%$ & 2.83  &  $75.84\%$ \tiny{\textcolor{red}{(+0.84\%)}} & $68.72\%$ \tiny{\textcolor{red}{(+5.51\%)}} &  $2.90$ \tiny{\textcolor{red}{(+1.40\%)}} & {2.58\%} \\ \hline
{Qwen 2.5-VL-7B} &$66.73\%$ & $61.21\%$ &  $1.30$ &  $67.76\%$ \tiny{\textcolor{red}{(+1.03\%)}} & $63.87\%$ \tiny{\textcolor{red}{(+2.66\%)}} & $2.46$ \tiny{\textcolor{red}{(+23.20\%)}}& {8.96\%} \\ \hline 
{Qwen 2.5-VL-32B} & $77.71\%$ & $67.50\%$ & $2.79$  &  $79.21\%$ \tiny{\textcolor{red}{(+1.50\%)}} & $70.24\%$ \tiny{\textcolor{red}{(+2.74\%)}} & $2.87$ \tiny{\textcolor{red}{(+1.60\%)}}&{1.95\%} \\ \hline 

\hline
\end{tabular}
\end{center}
\setlength{\abovecaptionskip}{0pt}
\end{table*}

\paragraph{Prompt construction and VLM inference.}
Once the example index $I^\star$ has been chosen, we obtain the corresponding description $t_{I^\star}$ and build a retrieval-augmented prompt: $\mathrm{Prompt}(x,q,t_{I^\star})
   = \textsc{Format}\big(q, t_{I^\star}\big)$,
where $\textsc{Format}(\cdot)$ is a fixed template that first presents the
example description and then asks the model to comment on the quality of the query CG image.
The query image $x$ and the textual prompt are fed into the VLM assistant,
which outputs both scalar quality judgments (e.g., ``very good'') and
free-form explanations.

\section{Experiments}
\subsection{Experiment settings}
We evaluate R4-CGQA on the validation set for ablation studies and on the testing set to assess state-of-the-art VLMs. The base set serves as the retrieval library.
For the multiple-choice (Choice) and yes-or-no questions, we directly compute accuracy using the ground-truth answers.
For the normal Q\&A (Q\&A), we use GPT-4o-mini to score the VLM outputs by comparing them with the ground-truth answers within a specific template.
Some VLMs (LLava, Llama 3.2-V, Bakllava, Gemma3 and MiniCPM-V) are accessed through the popular Ollama tool \cite{ollama}, whereas others are evaluated on an NVIDIA A800 80 GB GPU using PyTorch.
We denote the number of nearest neighbors in Eq.~(\ref{eq:topk}) by $K$ and the similarity threshold by $T$.
Additional results and details are in the supplementary materials.

\subsection{Evaluation on VLMs}
\label{sec:VLM_result}
As shown in Table ~\ref{fulltable}, we evaluate our method on VLMs, which involves LLava1.6 \cite{DBLP:conf/nips/LiuLWL23a}, Llama3.2-Vision \cite{meta_llama32_vision}, Gemma3 \cite{Kamath2025Gemma3}, Mini-CPM \cite{Yao2024MiniCPMV}, Bakllava \cite{BakLLaVA2023}, Pixtral \cite{Agrawal2024Pixtral12B}, LlavaNext \cite{Liu2024LLaVANeXT,Li2024LLaVAOneVision}, and Qwen 2.5-VL\cite{DBLP:journals/corr/abs-2502-13923}, on three types of questions in R4-CGQA: Choice, Yes-or-no, and Q\&A. R4-CGQA consistently improves every model on every reported metric. On the choice questions, the average gain is 4.26\% absolute, with Bakllava improving from 43.72\% to 55.97\% (+12.25 \%), and Mini-cpm and LLava-1.6-13B gaining +7.58\% and +7.47\%, respectively. On yes-or-no questions, the average improvement rises to 6.94\% absolute, and models such as Gemma3-4B and Llama~3.2-Vision increase by 11.67\% and 10.39\%. For the nine models with Q\&A scores, R4-CGQA yields an average increase of 0.32\% on the 5-point scale (6.40\% of the full score), with Gemma3-4B improving from 1.05\% to 2.32\%. Results reveal that our method is more capable of unlocking the potential of relatively small models in CGQA. It also indicate that R4-CGQA provides consistent and non-trivial gains for both weaker and stronger VLMs.

\begin{table} [t]
\small
\begin{center}
\caption{\label{tb:ab1} Ablation study of content and quality based retrieval. ``w/o." denotes without.}
\setlength{\tabcolsep}{0.5 mm}
\begin{tabular}{c|ccc|ccc}
\hline
   & \multicolumn{3}{c|}{Llava-1.6-7B}  &  \multicolumn{3}{c}{Llama3.2-V-11B}  \\
 Config. &  \tabincell{c}{Choice}& \tabincell{c}{Yes-or-no} &  \tabincell{c}{Q\&A} & \tabincell{c}{Choice}& \tabincell{c}{Yes-or-no} & \tabincell{c}{Q\&A}\\ 
 \hline \hline
Base   &$50.1\%$&$48.8\%$ &$2.23$ &$65.3\%$ &$55.8\%$ &$1.94$  \\
w/o. quality   &$56.8\%$&$59.0\%$ &$2.31$  &$61.0\%$ &$68.3\%$ &$2.21$  \\
w/o. content     &$57.0\%$ &$\textbf{60.4\%}$ &$2.29$  &$65.2\%$ &$68.9\%$ &$2.30$   \\ \hline
Full     &$\textbf{59.8\%}$ &$59.9\%$  &$\textbf{2.38}$   &$\textbf{66.7\%}$ &$\textbf{69.0\%}$  &$\textbf{2.43}$  \\
 \hline 
\end{tabular}
\end{center}
\end{table}

\begin{table} [t]
\small
\begin{center}
\caption{\label{tab:multiimage} Explore multi-image performance on Pixtral-12B; $T=0.9$ and $K=7$.}
\setlength{\tabcolsep}{0.9 mm}
\begin{tabular}{l|c|c|c|c}
\hline
Question & Base &  \tabincell{c}{Multi-image \\ (only)}  & \tabincell{c}{Multi-image \\ + R4-CGQA}  & {R4-CGQA}   \\
\hline \hline
Choice & $70.8\%$ & $68.5\%$ & $71.0\%$  & $73.5\%$  \\
Yes-or-no & $68.3\%$ & $71.6\%$ & $71.1\%$  & $71.7\%$  \\
\hline
\end{tabular}
\end{center}
\end{table}

\begin{table} [t]
\small
\begin{center}
\caption{\label{tab:Kexp} Explore the impact of different K values on the performance of Llava-1.6-7B responses; $T$ is set to $0.8$.}
\setlength{\tabcolsep}{1.4 mm}
\begin{tabular}{l|c|c|c|c|c}
\hline
Question & $K=1$ & $K=3$ &  $K=5$  & $K=7$   & $K=9$   \\
\hline \hline
Choice & $57.1\%$ & $ 59.0\%$ & $59.8\%$  & $58.7\%$ & $57.2\%$ \\
Yes-or-no & $57.5\%$ & $ 58.0\%$ & $59.9\%$  & $56.5\%$ & $57.0\%$ \\
\hline
\end{tabular}
\end{center}
\end{table}

\subsection{Ablation Study}
\label{sec:ablation}
\paragraph{Investigating content and quality retrieval.}
As shown in Table~\ref{tb:ab1}, ``w/o. quality" means removing the quality embedding module during similarity search, and ``w/o. quality"  denotes using REIQA rather than CLIP for content embedding and the quality embedding module is also removed.
For Llava-7B, the proposed full pipeline further improves the Choice accuracy to
$59.8\%$ and keeps a high Yes-or-no accuracy ($59.9\%$), yielding absolute
gains of $+9.7\%$ and $+11.1\%$ over the Base model. 
For Llama3.2-V, the same trend holds: quality-only retrieval is already
strong on Yes-or-no questions, but the full framework still achieves the best overall
results ($66.7\%$ / $69.0\%$ on Choice / Yes-or-no), with a large gain of
$+13.2\%$ on Yes-or-no accuracy. Besides, all of full pipelines achieve highest $Q\&A$ scores.
These results confirm that the content and quality based retrieval is more robust than using either branch alone.

\paragraph{Investigating multi-image input.} A natural question is why we do not simply input multiple CG images into the VLM at the same time.
In general, conventional VLMs do not perform robust comparative analysis over multiple images, so feeding multiple images often leads to a significant decline in performance. Nevertheless, we also conduct an experiment on Pixtral, which is capable of processing multiple images, as shown in Table~\ref{tab:multiimage}. The results show that when we input similar images together with the current image (Multi-image (only)), performance on the Choice metric decreases by $2.3\%$. Even when multiple images are combined with our method (Multi-image + R4-CGQA), performance on Choice is still $2.5\%$ lower than R4-CGQA alone. The results suggest that simply feeding multiple images into VLMs is not a reasonable solution for CGQA.

\begin{figure}[t]
  \centering
  \includegraphics[width=1.0 \linewidth]{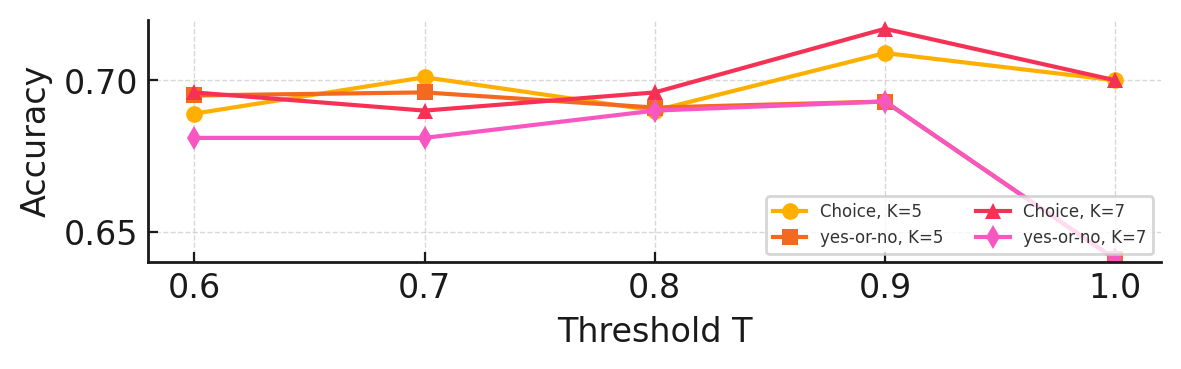}
  \caption{The influence of threshold $T$ and the accuracy rate of responses was tested on LlavaNext-8B, and we investigate $K=5$ and $K=7$. \label{fig:accuracy_vs_T_yesno} }
\end{figure}

\paragraph{Investigating $K$.}
Table ~\ref{tab:Kexp} shows the effect of the candidate set size $K$ for
Llava-7B with a fixed similarity threshold $T=0.8$.
Accuracy improves when increasing $K$ from $1$ to $5$ (up to $59.8\%$ /
$59.9\%$ on Choice / Yes-or-no), but drops again when $K$ is further
enlarged, indicating that a moderate number of neighbours is preferable to
either a single example or an overly large, noisy candidate set.
\paragraph{Investigating $T$.}
Figure ~\ref{fig:accuracy_vs_T_yesno} shows the impact of the threshold $T$ for
LlavaNext-8B and $K\in\{5,7\}$: accuracies are relatively stable when
$T$ lies between $0.7$ and $0.9$, and the configuration with $K=5$ is
consistently more stable than $K=7$. When $T=1.0$, i.e., do not choose similar image descriptions as examples, the performance decline drastically on Yes-or-no.

\section{Conclusion}
In this paper, we address the limitations of current CG quality assessment by introducing a new dataset of 3.5K CG images with detailed text-based quality descriptions covering six perceptual dimensions. Building upon this resource, we further propose a retrieval-augmented, two-stream framework that leverages both content similarity and quality similarity to supply VLMs with relevant example descriptions during inference, named R4-CGQA. Extensive experiments on several representative VLMs demonstrate that the retrieved textual cues substantially enhance their ability to provide accurate and interpretable CG quality judgments. Our contributions offer a scalable and training-free solution for empowering VLMs in CGQA, which can strongly support future research in this direction.


\bibliographystyle{ieeenat_fullname.bst}
\small
\bibliography{main.bib}
\end{document}